# Deep Spatial Regression Model for Image Crowd Counting

Haiyan Yao[1,2,3], Kang Han[1,2], Wanggen Wan[1,2], Li Hou[1,2]

[1]School of Communication and Information Engineering, Shanghai University, Shanghai, China

[2]Institute of Smart City, Shanghai University, Shanghai, China

[3]Anyang Institute of Technology, Henan, China

E-mail: yaohywqh@shu.edu.cn

**Abstract**

Computer vision techniques have been used to produce accurate and generic crowd count estimators in recent years. Due to severe occlusions, appearance variations, perspective distortions and illumination conditions, crowd counting is a very challenging task. To this end, we propose a deep spatial regression model(DSRM) for counting the number of individuals present in a still image with arbitrary perspective and arbitrary resolution. Our proposed model is based on Convolutional Neural Network (CNN) and long short term memory (LSTM). First, we put the images into a pretrained CNN to extract a set of high-level features. Then the features in adjacent regions are used to regress the local counts with a LSTM structure which takes the spatial information into consideration. The final global count is obtained by a sum of the local patches. We apply our framework on several challenging crowd counting datasets, and the experiment results illustrate that our method on the crowd counting and density estimation problem outperforms state-of-the-art methods in terms of reliability and effectiveness.

**Keywords:** crowd counting, convolutional neural network, long short term memory (LSTM), spatial regression.

## 1. Introduction

In the recent years, along with the increasing degree of urbanization, more and more people choose to live in the city. The benefits of this trend are enriching the cultural life and making full use of the convenient urban infrastructure. At the same time, a large scale of people gathers together to organize various activities, such as Olympic Games, religious rally, festival celebration, strike, Marathon, concert and so on. When tens of thousands of people gathering together in limited space, a tragedy is probably to happen. In Shanghai Bund, the new year's eve of 2015, 36 persons were killed and 49 persons were injured in a massive stampede. In order to avoid such deadly accidents, the research on automatic detection and counting and density in large scale crowd is playing a significant role in city security and city management.

In computer vision, many studies have focused on how to establish models which can accurately estimate the numbers of pedestrians in images and videos. These models can be extended to be applied on other domains, such as vehicles estimation at traffic junctions or super highway [3], animal crowd estimation in wildlife migration, quantification of specific populations of cells for precision diagnostic in laboratory medicine [1].

The challenges in the crowd counting and density estimation are the severe occlusions, appearance variations, perspective distortions and illumination conditions which affect the performance of the model in different degrees. Specifically, the density and distribution of crowd vary significantly in the crowd counting task. This phenomenon can be observed in the datasets we can access. Figure 1 illustrates some examples of the datasets for our experiments. To tackle these challenges, we propose a new framework for counting the number of individuals present in a still image with arbitrary perspective and arbitrary resolution. First, we put the images into a pre-trained CNN to extract a set of high-level features. Then the features in adjacent regions are used to regress the local counts with long short term memory (LSTM) [8] structure which takes the spatial information into consideration. The final global count is obtained by a sum of the local patches. Our approach achieves the state-of-the-art results on all of these challenging datasets and demonstrate the effectiveness and reliability.

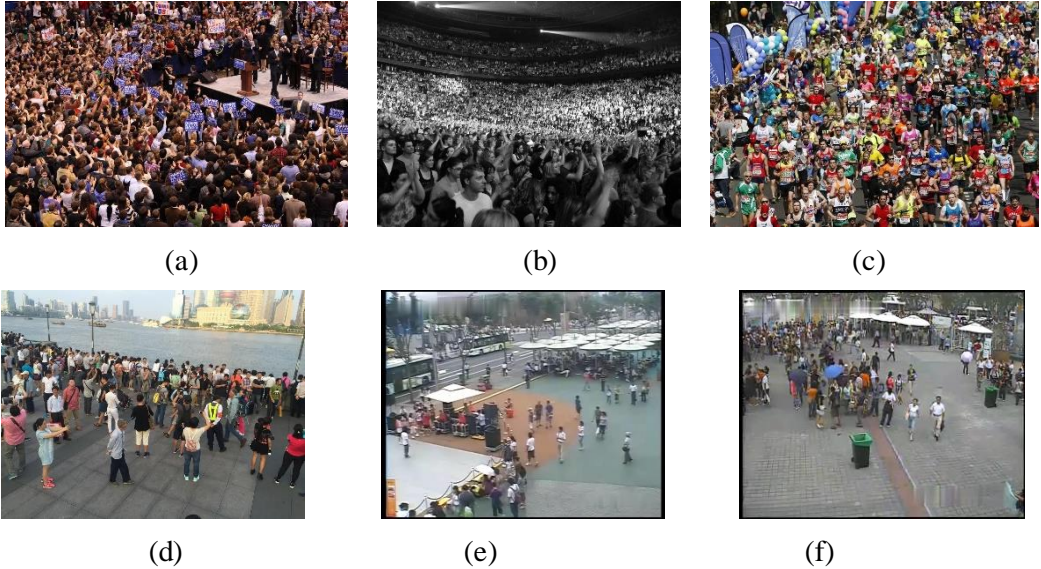

(a) (b) (c)

(d) (e) (f)

Figure 1. Samples of different datasets. (a) Example images of the Shanghaitech dataset Part_A. (b) Example grey scale images of the UCF_CC_50 dataset. (c) Example images of the AHU-CROWD dataset. (d) Example images of the Shanghaitech dataset Part_B (e)(f)Example frames of the WorldExpo'10 dataset.

The main contributions of our study are as follows: We propose a deep spatial regression model to estimate the people counting in images. Due to the variability of camera view-point and density, strong correlation exists in transverse direction. Overlapping regions strategy makes this correlation stronger. A novel deep features matrix is set up which contains the spatial information. Our deep spatial regression model can learn the spatial constraint relation of local counts in adjacent regions effectively and improve the accuracy significantly.

The rest of this paper is organized as follows. In Section 2, we briefly review the related work of crowd counting and density estimation. Then a novel DSRM estimation model is proposed in Section 3. Experimental results for our proposed framework on datasets of different density distribution are presented in Section 4. Finally, conclusions are drawn in Section 5.

**2. Related work**

Many studies have been made in the literature of crowd counting. In the early days, many methods adopted a counting-by-detection strategy. [9] has used a count estimation method which combines foreground segmentation and head-shoulder detection. They first detect active areas, then detect heads and count the number from the foreground areas. Cheriyadat et al. [10] have proposed an object detection system based on coherent motion region detection for counting and locating objects in the presence of high object density and inter-object occlusions. [11] has utilized an unsupervised data-driven Bayesian clustering algorithm which detect the individual entities. These counting-by-detection methods attempt to determine the number of people by detecting individual entities and their locations simultaneously. However, the performance of the detectors reduces dramatically when dealing with dense crowds and severe occlusion. Most of these works experiment on datasets containing sparse crowd scenes, such as UCSD dataset [12], Mall dataset[13] and PETS dataset[14].

Loy et al. [15] have proposed semi-supervised regression and data transferring approaches to reduce the amount of training set. Further work by Idrees et al. [20] has presented a method which combines different kinds of hand-crafted features, i.e. HOG based head detections, Fourier analysis, and interest points based counting. Once they estimate density and counts in each patch by combined features, they place them in multi-scale Markov Random Field to smooth the results in counts among nearby patches. Although this method relatively improves the accuracy, it is still dependent on traditional hand-engineered representations, e.g. SIFT [4], HOG [5], LBP [6].

In recent years, deep learning has attracted people's attention. Some studies [7, 16] have shown that the features extracted from deep models are more effective than hand-crafted features for many applications. For example, methods of deep learning have remarkably improved the state-of-the-art in visual object recognition, speech recognition, object detection and many other domains [17]. In order to adapt the change of the crowd density and perspective, [22] has introduced a multi-column CNN (MCNN) model to estimate the density map of a still image. Each column has filters with receptive fields of different sizes. They pre-train each single column separately and then fine-tune the multi-column CNN. Zhang et al.[21] have proposed a CNN model of iterative switchable training scheme with two objectives: estimating global count and density map. Firstly they pre-train their CNN model based on all training set. Then they retrieve the samples with the similar distribution to the test scene and added them to the training data to fine-tune the CNN model. Perspective maps of frames are used in this process which can significantly improve the performance. Unfortunately, generating perspective maps on both training scenes and test scenes is computationally complex and time-consuming, which limits the applicability of this method. Generally speaking, these neural networks contain less than seven layers.

Currently, many deep neural networks produce amazing results on classification, object detection, localization and segmentation tasks. Several attempts have been made to apply these deep models to crowd counting and density estimation. Boominatahn et al. [24] used a combination of deep (VGG-16 [31]) and shallow fully convolutional networks to predict the density map for a dense image. They evaluated their approach on only one dataset, but the experiment result is not competitive. Shang et al.[23] introduced an end-to-end CNN network that directly maps the whole image to the counting result. A pre-trained GoogLeNet model [32] is used to extract high-level deep features and the LSTM decoders for the local count and fully connected layers for the final count. The authors resize images to 640×480 pixels before feeding them to the network, which will bring the errors.

Our approach is related to the ResNet model [2], which is trained on ImageNet dataset and gets the perfect score on the classification task. The 152 layer ResNet is utilized to extract deep features from the patches cropped from the whole images with overlaps. Due to the 50% overlap, the crowd counts of the adjacent patches have high correlation. So we use a LSTM structure considering the spatial information to regress the local counts. Finally, the total number of a still image is the sum of the local counts.

## 3 Method
### 3.1 System overview

In this section, we give a general overview of the proposed method, details are provided in the following sections. In this paper, we propose a deep spatial regression model for crowd counting and density estimation which is shown in Figure 2.

In the first place, we feed the patches cropped from the whole image to a pre-trained CNN called ResNet. The 152-layer residual net is the deepest network ever presented on ImageNet and still has lower complexity than VGG nets. The purpose is to get the 1000 dimensional high-level features. We crop 100×100 patches with 50% overlap from every image. This data augmentation helps us to address the problem of the limited training set. Meanwhile, due to the irregularity of the crowd distribution and non-uniform in large scale, smaller slices can make it to be approximately uniform distribution.

Moreover, in order to get the accurate local counts, a novel deep features matrix which contains the features extracted by the ResNet is learned by a LSTM neural network. In this process, the spatial constraint relation of local counts in adjacent regions is considered to improve the accuracy of the estimated result.

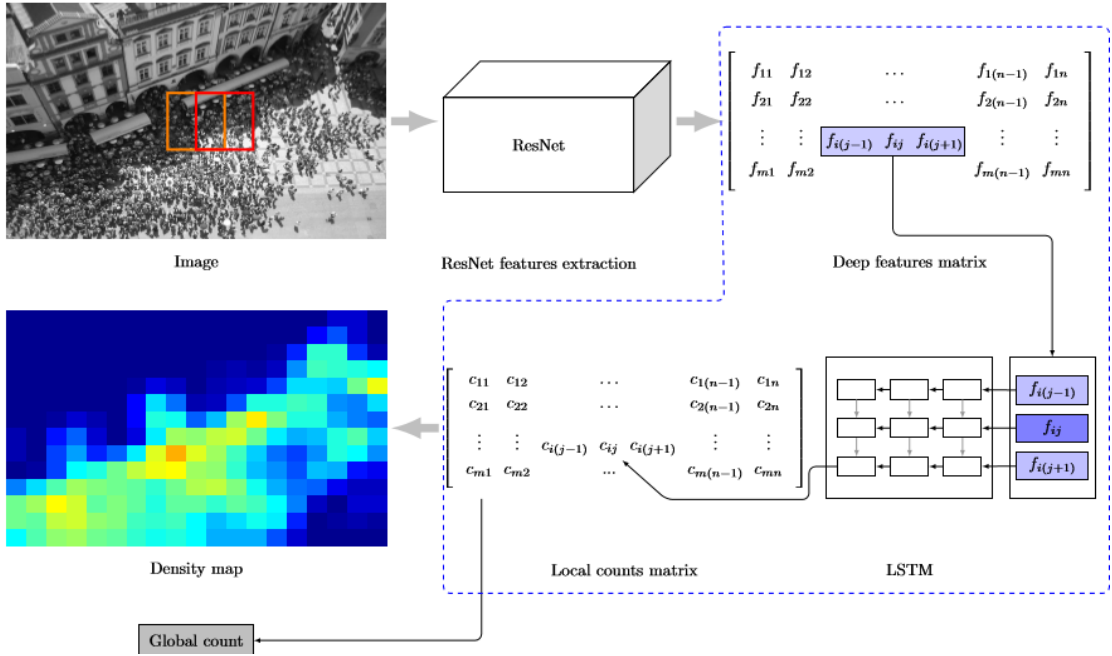

Figure 2. Overview of our proposed DSRM crowd counting method. In the dashed box, the block diagram of the LSTM structure considering the spatial information to regress the counts is shown.

Finally, we obtain the local counts matrix consisting of local counts in every patch. The final count of the whole image is the sum of the local counts. Furthermore, we get the more intuitive density map which can be seen clearly in Figure 2.

The Euclidean distance is used to measure the difference between the ground truth and the prediction count. The loss function is defined as follows.

$$L(\theta) = \frac{1}{N}\sum_{i=1}^{N}\|F(X_i;\theta) - z_i\|^2 \qquad (1)$$

where N is the number of the training image patches in the dataset and $\theta$ is the parameters of the framework. $L(\theta)$ is the loss between the regressed count $F(X_i;\theta)$ from the network and the ground truth $z_i$ of the image patches $X_i$ ( i = 1,2,…,N). The loss is minimized using mini-batch gradient descent and back-propagation.

### 3.2 LSTM regression with spatial information

The spatial information in a given image is important besides the crowd counts. Since the local patches have overlapping portions, the crowd counts of the adjacent patches have a high correlation. We obtain much more accurate global counts through estimation of the counting in local patches. In the work of [13] makes use of the correlation and Markov random field to smooth the counting results of the local patches. But this method does not take advantage of the spatial information between the adjacent patches during training. The spatial information is important for final counting. Some studies utilize density maps which can preserve more distribution information to regress the counting number [21，22].

We serialize the data of local patches and the spatial relation is learnt by recurrent neural network (RNN). In general, the RNN is used to process temporal sequences. We extend it to the spatial sequences. RNN has been applied in many fields, such as language modeling, image captioning, speech recognition and location prediction. LSTM is a special kind of RNN architecture, capable of learning long-term dependencies. It has achieved considerable success and has been widely used. LSTMs [8] are explicitly designed to avoid the long-term dependency problem.

In our proposed framework, LSTM is used to learn the spatial constraint relation of local counts in adjacent regions, so that more accurate estimation results can be obtained by regression. The dashed box in Figure 2 shows the block diagram of our proposed local counts regression system. It contains three parts: the input part is the deep features matrix, the core part is the local counts regression network based on LSTM and the output part is the global counts of a single image.

The deep features matrix in which the local features are arranged according to the spatial relation of the image patches. For each local patch split from a single image, a deep ResNet is used to extract the high-level features representing the local number of individuals. Through our careful observation of the crowd images in all of the datasets considered in this study, we deduce that in a single crowd image, the sizes of heads are strongly related in transverse direction, whereas in the longitudinal direction the correlation is very poor due to the angle view, illumination condition and occlusion. Therefore, we consider the horizontal serial of the local features matrix. The details of applying our method can be seen in Figure 2. For any local feature, the adjacent two set of features (left and right sides) are used to form space series with the length of three. In order to make the features of edges also form spatial sequence data, we extend the local part by copying the marginal columns of deep features matrix. With regard to a local features matrix $F$ of a single crowd image, the spatial sequence $x_{ij}$ whose length is 3 can be expressed as follows:

$$x_{ij} = \begin{cases} (f_{i1}, f_{i1}, f_{i2}) & j = 1 \\ (f_{i(j-1)}, f_{ij}, f_{i(j+1)}) & j \in [2, n-1] \\ (f_{i(n-1)}, f_{in}, f_{in}) & j = n \end{cases} \quad (2)$$

where i and j are the row and column indexes of the features matrix. In the features matrix, m and n in $F$ represent the number of the rows and columns of the local crowd image features respectively.

The LSTM neural network has three layers. The first and second layers contain 100 neurons and the size of output spatial sequence is $3 \times 100$. The last layer has only one neuron and outputs the regression local count $c_{ij}$. We use Adam algorithm [33] in training process.

## 4. Experiment

We evaluate our algorithm on four crowd counting datasets. Comparing to most CNN based methods in the literature, the proposed DSRM model achieved excellent performance in all the datasets. Implementation of the proposed model and its training are based on the TensorFlow deep learning framework, using NVIDIA Tesla K20 GPU.

Mean absolute error (MAE), mean squared error (MSE) and mean normalized absolute error (MNAE) are utilized to evaluate and compare the performance of different methods. These three metrics are defined as follows:

$$\text{MAE} = \frac{1}{M} \sum_{i=1}^{M} |z_i - \hat{z}_i| \quad (3)$$

$$\text{MSE} = \sqrt{\frac{1}{M} \sum_{i=1}^{M} (z_i - \hat{z}_i)^2} \quad (4)$$

$$\text{MNAE} = \frac{1}{M} \sum_{i=1}^{M} \frac{|z_i - \hat{z}_i|}{z_i} \quad (5)$$

where M is the number of images in the test set, $z_i$ is the ground truth of people in the $i_{th}$ image and $\hat{z}_i$ is the prediction value of people in the $i_{th}$ image. MAE indicates the accuracy of the estimation, the MNAE is related to the MAE which represents the average deviation rate. The MSE indicates the robustness of the estimates. Lower MAE, MNAE and MSE values mean more accuracy and better estimates.

### 4.1 Experiment on the Shanghaitech dataset

The Shanghaitech dataset includes 1198 images with about 330,165 humans with centers of their heads annotated. This is a large scale crowd counting dataset. It is split into two parts: Part_A contains 482 high-density images (up to 3000 people) which are crawled from the Internet, Part_B includes 716 medium density images (up to 600 people) which are taken from the streets of Shanghai. The resolution of images contained in Part_A is different while the resolution in Part_B is the same 768×1024 pixels. By following the same setting as in [22], we use 300 images as training set in Part_A and 400 images in Part_B.

We compare our approach with existing methods, including traditional hand-crafted feature approach (LBP+RR) and the density map regression approach (MCNN). The results of our proposed method DSRM are shown in the last row of Table 1. Our DSRM model can estimate crowd counts effectively. Some of our results including counting results and density maps are shown in Figure 3. In more detail, we compare the estimated count with the ground truth by line graph. According to crowd counts in ascending order, the images of Shanghaitech Part_A and Part_B are divided into 10 groups respectively. The comparison result is shown in Figure 4.

Table 1. Comparing results on Shanghaitech dataset using MAE and MSE

| Method | Part_A | | Part_B | |
|---|---|---|---|---|
| | MAE | MSE | MAE | MSE |
| LBP+RR [22] | 303.2 | 371.0 | 59.1 | 81.7 |
| Zhang et al. [21] | 181.8 | 277.7 | 32.0 | 49.8 |
| MCNN [22] | 110.2 | 173.2 | 26.4 | 41.3 |
| CNN-MRF[13] | 79.1 | 130.1 | 17.8 | **26.0** |
| DSRM | **74.4** | **114.7** | **15.2** | 29.0 |

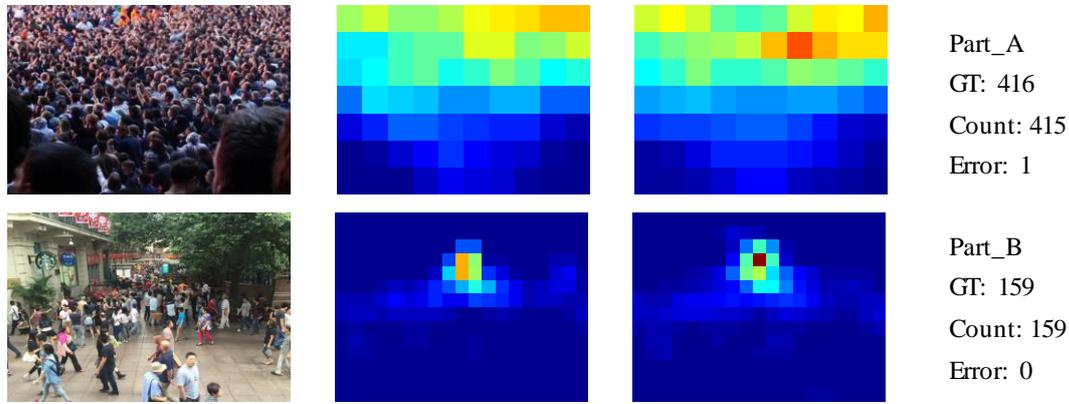

Figure 3. Our counting results and density maps on the Shanghaitech Part_A, Shanghaitech Part_B datasets. (Left) sample selected from each test scene. (Middle) ground truth density map on the sample. (Right) estimated density map on the sample. The ground truth, estimated count and absolute error are shown at the right of the maps.

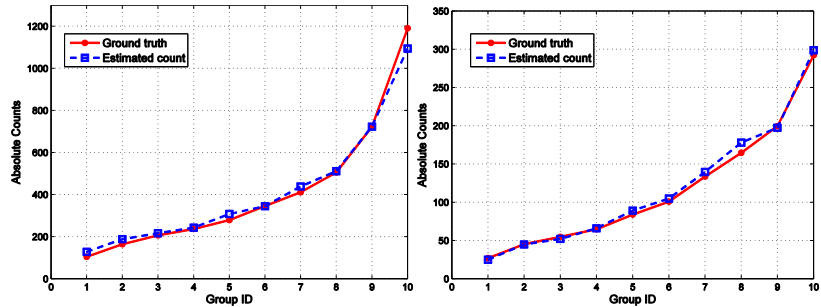

Figure 4. The comparison of the ground truth and the estimated count on Shanghaitech Part_A dataset (left) and Shanghaitech Part_B dataset (right). Absolute counts in the vertical axis is the average crowd number of images in each group.

**4.2 Experiment on the UCF_CC_50 dataset**

The UCF_CC_50 dataset is a very challenging dataset introduced by Idrees et al. [20]. There are only 50 gray scale images taken from the Internet. The numbers of people are ranging from 94 to 4543 with an average of 1280 pedestrians in each image.

Following the convention we perform 5-fold cross validation on this dataset, MAE and MSE are employed as the evaluated metrics. This dataset is a classical benchmark, so we can compare the performance of different methods.

Rodriguez et al. [18] make use of a density map estimation and the location of individual people to get better head detection results in crowded video scenes. Lempitsky et al. [19] introduce a new loss function called MESA distance to learn a density regression model using dense SIFT features on randomly selected patches. Idrees et al. [20] combine three hand-crafted features to estimate density and counts in each patch and place them in multi-scale Markov Random Field to smooth the results in counts among nearby patches. Zhang et al. [21] propose a CNN model of a switchable training scheme with two tasks: estimating global count and density map. Perspective maps of frames are used in the process. Zhang et al. [22] introduce a multi-column CNN (MCNN) model to estimate the density map of a still image. Each column has filters with receptive fields of different sizes.

Table 2. Comparison results on the UCF_CC_50 dataset

| Method | MAE | MSE |
|---|---|---|
| Rodriguez et al. [18] | 655.7 | 697.8 |
| Lempitsky et al. [19] | 493.4 | 487.1 |
| Idrees et al.[20] | 419.5 | 541.6 |
| Zhang et al. [21] | 467.0 | 498.5 |
| MCNN [22] | 295.1 | 490.2 |
| **DSRM** | **283** | **372** |

Since the image of the UCF dataset is gray, we extend the image to three channels by copying the data. We compare our proposed methods with five existing approaches on the UCF_CC_50 dataset in Table 2. Our proposed method achieved the best MAE and MSE. The counting results and density maps selected from our experiment can be seen in Figure 5. Similar to Figure 4, the comparison of the ground truth and the estimated count is illustrated in Figure 6.

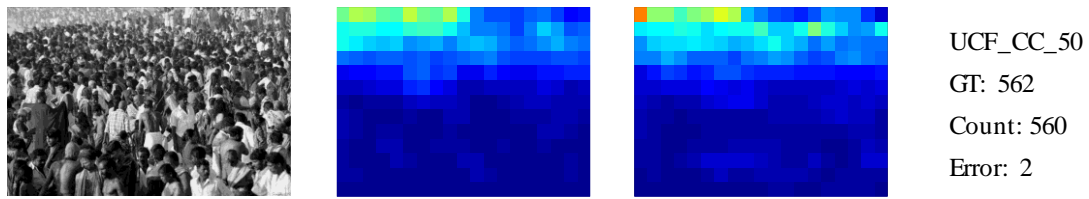

UCF_CC_50
GT: 562
Count: 560
Error: 2

Figure 5. Our counting results and density maps on the UCF_CC_50 datasets. (Left) sample selected from each test scene. (Middle) ground truth density map on the sample. (Right) estimated density map on the sample. The ground truth, estimated count and absolute error are shown at the right of the maps.

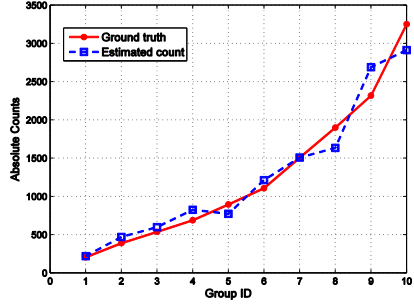

Figure 6. The comparison of the ground truth and the estimated count on UCF_CC_50 dataset. Absolute counts in the vertical axis is the average crowd number of images in each group.

**4.3 Experiment on the AHU-CROWD dataset**

We also evaluate our method on the AHU-CROWD dataset. This dataset contains 107 crowd images with 45,000 annotated human totally. The count of pedestrians is ranging from 58 to 2201. Similar with the UCF_CC_5 dataset, we perform 5-fold cross validation.

Hu et al. [26] extracts deep crowd features through a ConvNet structure. Then two signals, i.e. crowd counts and crowd density are used to regress the counting in the local region. In addition, they applied density level (mid-level and high-level) as density label which is convenient in the practical situation. In order to compare with the result of [26], we use the same evaluation criteria, e.g. MAE and MNAE. Our further results of MSE are listed in the last column of Table 3. We can see that our approach is highly superior to the current best method.

In Figure 7, a sample of AHU-CROWD dataset and the corresponding results and density maps are shown. A similar comparison of ground truth and estimated counts can be seen in Figure 8.

Table 3. Comparison results of different methods on the AHU-CROWD dataset

| Method | MAE | MNAE | MSE |
|---|---|---|---|
| Haar Wavelet [27] | 409.0 | 0.912 | - |
| DPM [28] | 395.4 | 0.864 | - |
| BOW–SVM [29] | 218.8 | 0.604 | - |
| Ridge Regression [30] | 207.4 | 0.578 | - |
| Hu et al.[26] | 137 | 0.365 | - |
| **DSRM** | **81** | **0.199** | **129** |

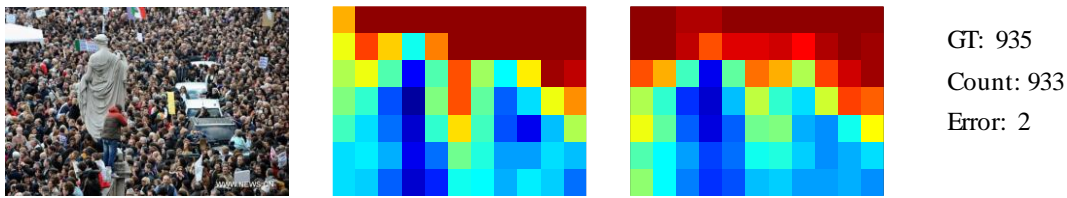

Figure 7. Our counting results and density maps on the AHU-CROWD datasets. (Left) sample selected from each test scene. (Middle) ground truth density map on the sample. (Right) estimated density map on the sample. The ground truth, estimated count and absolute error are shown at the right of the maps.

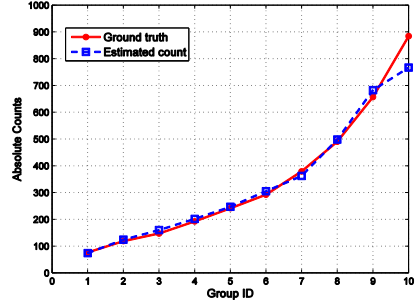

Figure 8. The comparison of the ground truth and the estimated count on AHU-CROWD dataset. Absolute counts in the vertical axis is the average crowd number of images in each group.

**4.4 Experiment on the WorldExpo'10 dataset**

WorldExpo'10 crowd counting dataset was introduced by Zhang et al.[21]. This dataset consists of 1132 annotated video sequences which are captured by 108 surveillance cameras, all from 2010 WorldExpo. The authors labeled a total of 199,923 pedestrians at the centers of their heads in 3,980 frames. The resolution is uniformly obtained by 576×1024 pixels. The dataset is split into two parts. 3,380 images from 103 scenes are treated as the training set. The testing set has 5 different video sequences and 120 labeled frames in each one. The pedestrian number of the dataset range from 1 to 253.

Table 4. MAE of the WorldExpo'10 crowd counting dataset

| Method | Scene 1 | Scene 2 | Scene 3 | Scene 4 | Scene 5 | Average |
|---|---|---|---|---|---|---|
| Luca Fiaschi et al.[25] | 2.2 | 87.3 | 22.2 | 16.4 | 5.4 | 26.7 |
| Ke et al.[30] | 2.1 | 55.9 | **9.6** | 11.3 | 3.4 | 16.5 |
| Zhang et al.[21] | **2.0** | 29.5 | 9.7 | **9.3** | **3.1** | 10.7 |
| DSRM | 3.0 | **10.3** | 9.8 | 14.4 | 4.8 | **8.4** |

For fair comparison, our experiment follows the work of [21]. In this dataset, we feed half of the test set in each scene into our network as training set and test on the remaining frames. Zhang et al.[21] use different methods to compare the performances of five scenes. The mean absolute errors are good in Scene 1, Scene 3, Scene 4 and Scene 5. However, for Scene 2, a worse result is achieved due to a large number of stationary crowds and cannot segment foreground accurately. Our algorithm overcomes this problem and gets the competitive results especially in challenging Scene 2. Note that our method does not rely on foreground segmentation and is tested on the whole image rather than just the area of ROI. We achieve the best average mean absolute error and comparable results in the five test scenes. Details can be seen in Table 4. The density estimation and counting results on the five test scenes are shown in Figure 9.

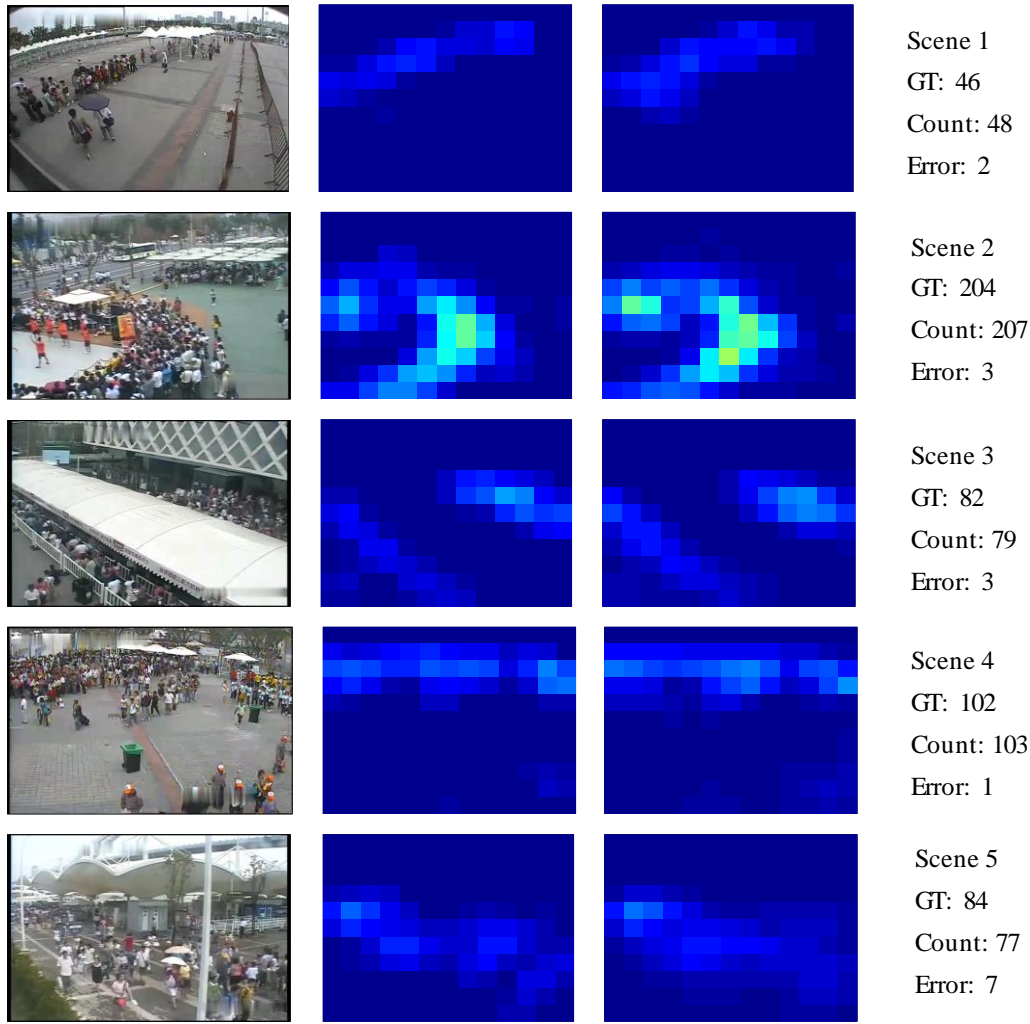

Figure 9. Our counting results and density maps on the WorldExpo'10 counting dataset. (Left) sample selected from each test scene. (Middle) ground truth density map on the sample. (Right) estimated density map on the sample. The ground truth, estimated count and absolute error are shown at the right of the maps.

**4.5 Evaluation on transfer learning**

In many studies, it is assumed that training data and testing data are taken from the same domain, such that the input feature space and data distribution characteristics are the same. Nevertheless, in numerous real-world applications, this assumption does not hold. Different research groups have proposed many datasets for crowd counting. Data collected by a research group might only include certain types of variations. For example, Shanghaitech Part_B is taken from the streets of metropolitan Shanghai. In the WorldExpo'10 dataset, video sequences are captured by surveillance cameras from Shanghai 2010 WorldExpo. Shanghaitech Part_A, UCF and AHU datasets are scrawled from the Internet. We can obviously see the distribution variation from histograms of crowd counts in Figure 10. The statistics of these different datasets are shown in Table 5.

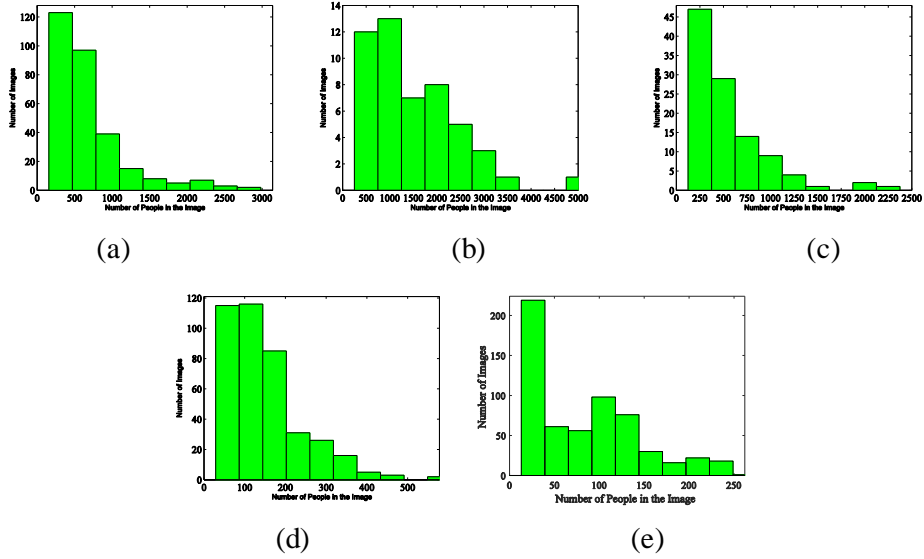

(a)                  (b)                 (c)

(d)                 (e)

Figure 10. Histograms of crowd counts of different datasets. (a) Histograms of Shanghaitech Part_A dataset, (b) Histograms of UCF_CC_50 dataset, (c) Histograms of AHU-CROWD dataset, (d) Histograms of Shanghaitech Part_B dataset (e) Histograms of test set of WorldExpo'10 dataset

Table 5. The statistics of different datasets. N is the number of images in the dataset; $N_{train}$ is the number of the train set, and $N_{test}$ is the number of the test set, $N_{people}$ is the count of people in the images, Max is the maximal count, Min is the minimal count and Average is the average count in the dataset. "5-fold" means 5-fold cross validation.

| Dataset | N | $N_{train}$ | $N_{test}$ | $N_{people}$ | | | Resolution |
|---|---|---|---|---|---|---|---|
| | | | | Max | Min | Average | |
| UCF_CC_50 | 50 | 5-fold | | 4543 | 94 | 1279.5 | different |
| Shanghaitech Part_A | 482 | 300 | 182 | 3139 | 33 | 501.4 | different |
| AHU-COWD | 107 | 5-fold | | 2201 | 58 | 420.6 | different |
| Shanghaitech Part_B | 716 | 400 | 316 | 578 | 9 | 123.6 | 768×1024 |
| WorldExpo'10 | 3974 | 3374 | 600 | 253 | 1 | 50.2 | 576×720 |

After analysis we can obtain the different distribution between these famous datasets concerning to crowd counting and estimation: Shanghaitech Part_A, AHU-CROWD and UCF_CC_50 have similar count distribution. Shanghaitech Part_B and WorldEXPO'10 have a similar distribution of pedestrian numbers. On the other hand, the total number of images in the datasets are different. We assume that using more training data will result in a better model. So we split these five datasets into source domain and target domain. The one which has the most number of images is chosen to be the source domain. The model is trained on the source dataset first, then transfer the learned model to the target dataset which helds a similar level of density.

To test and verify this idea, first we choose to train our algorithm on Shanghaitech Part_A, then the model was fine-tuned on the AHU-CROWD. The experiment results validate this hypothesis which is shown in Table 6. It can be seen that the MAE, MSE and MNAE are reduced in different degrees in AHU-CROWD. This surprising enhancement of accuracy also happens on the challenging UCF_CC_50 dataset which is illustrated in Table 7.

Table 6. Transfer learning on AHU

| Method | MAE | MNAE | MSE |
|---|---|---|---|
| DSRM | 81 | 0.199 | 129 |
| DSRM trained on Part_A | **77** | **0.179** | **128.03** |

Table 7. Transfer learning on UCF_CC_50

| Method | MAE | MNAE | MSE |
|---|---|---|---|
| DSRM | 283 | 0.288 | 372 |
| DSRM trained on Part_A | **240** | **0.223** | **350** |

Fuller experiment are conducted on the WorldExpo'10 crowd counting dataset. By fin-tuning the three layers of LSTM structure with training data in WorldExpo'10 crowd counting dataset, the accuracy can be greatly improved. The reason is that the knowledge of both the source and target data can be combined to improve the performance. Firstly, we fine-tune the model trained on Shanghaitech Part_A, the results are listed in Table 8 and Table 9. The two metrics MAE and MSE are used to evaluate the performance respectively. Secondly, we fine-tune the model trained on Shanghaitech Part_B, we can see the experiment results in the tables. It is clear that the second method is better than the first method both in MAE and MSE. The reason is the data distribution of Shanghaitech Part_B is similar to the target domain WorldExpo'10. This results validate our hypothesis and better accuracy is obtained.

Table 8. MAE of the WorldExpo'10 crowd counting dataset. ModelA means the model trained on Shanghaitech Part_A, ModelB means the model trained on Shanghaitech Part_B.

| Method | Scene 1 | Scene 2 | Scene 3 | Scene 4 | Scene 5 | Average |
|---|---|---|---|---|---|---|
| DSRM | **3.0** | 10.3 | 9.8 | 14.4 | 4.8 | 8.4 |
| Fine-tune the modelA | 4.2 | **9.8** | 11.1 | 8.8 | **3.4** | 7.46 |
| Fine-tune the modelB | 3.5 | 10.9 | **9.0** | **8.3** | 4.1 | **7.16** |

Table 9. MSE of the WorldExpo'10 crowd counting dataset

| Method | Scene 1 | Scene 2 | Scene 3 | Scene 4 | Scene 5 | Average |
|---|---|---|---|---|---|---|
| DSRM | 4.1 | 15.0 | 12.2 | 17.7 | 7.4 | 11.3 |
| Fine-tune the modelA | 5.5 | **14.5** | 13.5 | 10.9 | **5.4** | 9.96 |
| Fine-tune the modelB | **4.6** | 15.6 | **11.5** | 10.3 | 7.2 | **9.84** |

## 5. Conclusion

We have present a deep spatial regression model (DSRM) to estimate the counts of still images. Our general model is based on a Convolutional Neural Network (CNN) and long short term memory (LSTM) for crowd counting taking spatial information into consideration. With the overlapping patches divided strategy, the adjacent local counts are highly correlated. So we feed the images into a pre-trained convolutional neural network to extract high-level features. The features in adjacent regions are leveraged to regress the local counts with a LSTM structure considering the spatial information. Then the final global count of a single image is obtained by the sum of the local patches. We perform our approach on several challenging crowd counting datasets, and the experiment results illustrate that our deep spatial regression model outperforms state-of-the-art methods in terms of reliability and effectiveness.

**Acknowledgement**

This work is supported by the National Natural Science Foundation of China (61373084), Shanghai Science and Technology Committee Research Plan Project (17511106802).